# Adversarial Deformation Regularization for Training Image Registration Neural Networks

Yipeng Hu[1,2], Eli Gibson[1], Nooshin Ghavami[1], Ester Bonmati[1], Caroline M. Moore[3], Mark Emberton[3], Tom Vercauteren[1], J. Alison Noble[2], Dean C. Barratt[1]

[1] Centre for Medical Image Computing, University College London, London, UK
[2] Institute of Biomedical Engineering, University of Oxford, Oxford, UK
[3] Division of Surgery & Interventional Science, University College London, London, UK

**Abstract.** We describe an adversarial learning approach to constrain convolutional neural network training for image registration, replacing heuristic smoothness measures of displacement fields often used in these tasks. Using minimally-invasive prostate cancer intervention as an example application, we demonstrate the feasibility of utilizing biomechanical simulations to regularize a weakly-supervised anatomical-label-driven *registration network* for aligning pre-procedural magnetic resonance (MR) and 3D intra-procedural transrectal ultrasound (TRUS) images. A *discriminator network* is optimized to distinguish the registration-predicted displacement fields from the motion data simulated by finite element analysis. During training, the registration network simultaneously aims to maximize similarity between anatomical labels that drives image alignment and to minimize an adversarial *generator loss* that measures divergence between the predicted- and simulated deformation. The end-to-end trained network enables efficient and fully-automated registration that only requires an MR and TRUS image pair as input, without anatomical labels or simulated data during inference. 108 pairs of labelled MR and TRUS images from 76 prostate cancer patients and 71,500 nonlinear finite-element simulations from 143 different patients were used for this study. We show that, with *only* gland segmentation as training labels, the proposed method can help predict physically plausible deformation without any other smoothness penalty. Based on cross-validation experiments using 834 pairs of independent validation landmarks, the proposed adversarial-regularized registration achieved a target registration error of 6.3 mm that is significantly lower than those from several other regularization methods.

## 1 Introduction

The most recent image registration methods based on convolutional neural networks employ regularization strategies that incorporate non-application-specific prior knowledge of deformation between images to register. Unsupervised learning methods that maximize similarity measures between two images, e.g. [1, 2], rely on transformation parameterization via rigid or spline-based models, and/or smoothness penalty terms, such as the norm of displacement gradients, to predict physically plausible deformation. For supervised learning approaches, e.g. [3], deformation regularization is embedded in surrogate ground-truth displacements, such as those obtained from classical registration methods, to predict detailed voxel-level displacements.

For instance, anatomical labels have been proposed to drive a so-called weakly-supervised learning method to infer dense displacements for interventional multimodal image fusion applications [4], which commonly lack a robust intensity-based similarity measure and ground-truth deformation. For training their network, more than 4,000 anatomical structures were manually delineated from prostate cancer patient images. Obtaining sufficient anatomical landmarks is constrained not only by the substantial expert effort in labelling volumetric data, but also by inherent limitations on the number of available corresponding anatomical features from different imaging modalities (in this case MR and TRUS). In the same clinical application, using fewer anatomical labels for training leads to significantly larger target registration errors (TREs), whilst we show in this paper that deformation regularization is important to avoid overfitting to limited labels.

We further argue that application-specific biologically-plausible prior on organ motion may lessen the quantity and/or quality of anatomical labels required for training data-driven registration methods. Biomechanical finite-element (FE) simulations of intraoperative prostate motion, modelling nonlinear, anisotropic and inhomogeneous properties of soft tissue, have been applied to constrain pair-wise multimodal non-rigid image fusion [5-7]. In particular, population-based motion models from previous patient data that can be instantiated to provide patient-specific constraints for unseen data, e.g. [7], have advantages in the prostate modelling: FE simulations can be generated using MR images from patients whose TRUS images are not available and the registration network can be fine-tuned for imaging-protocol-specific data without repeating large numbers of simulations. However, fully-unsupervised generative modelling of complex biomechanical simulations over the entire deformation domain (as opposed to modelling only shapes or surfaces) is non-trivial and has not been applied to neural-network-based registration methods.

We demonstrate, to our knowledge for the first time, that it is feasible to optimize an end-to-end registration network using an adversarial strategy that penalizes the divergence between the registration-predicted deformation and the FE-simulated training data. The resulting automatic registration is useful to support a wide range of interventional real-time applications, such as focal therapy and targeted biopsy [8].

## 2 Method

### 2.1 Adversarial Deformation Regularization

During the training of a registration network, the network parameters $\theta^{(reg)}$ are optimized to predict a dense displacement field (DDF) that warps the moving image to spatially align with the fixed image, by minimizing a *registration loss* $\mathcal{L}^{(reg)}$. We propose a second neural network, the discriminator $D$ with parameters $\theta^{(dis)}$, which is simultaneously optimized to classify the registration-network-predicted DDF and the FE-simulated DDF by minimizing a *discriminator loss* $\mathcal{L}^{(dis)}$. Considering the registration network as a DDF generator in adversarial learning [9], the registration loss can be regularized by an additive *generator loss* $\mathcal{L}^{(gen)}$, weighted by a scalar hyper-parameter $\lambda_{adv}$. During every gradient-descent iteration, each of the two parameter sets $\theta^{(dis)}$ and $\theta^{(reg)}$ is updated once to minimize $\mathcal{L}^{(dis)}$ and $\left(\mathcal{L}^{(reg)} + \lambda_{adv} \cdot \mathcal{L}^{(gen)}\right)$, respectively, while the other set is kept fixed. In Section 2.2, we describe a registration loss for a weakly-supervised learning method (illustrated in Fig. 1 as the lighter shaded components) for registering prostate MR- and TRUS images. In Section 2.3, we introduce the discriminator- and generator losses for this application, which we show lead to stable and effective training of the proposed adversarial regularization. Details of the network architectures and their training are provided in Section 2.4.

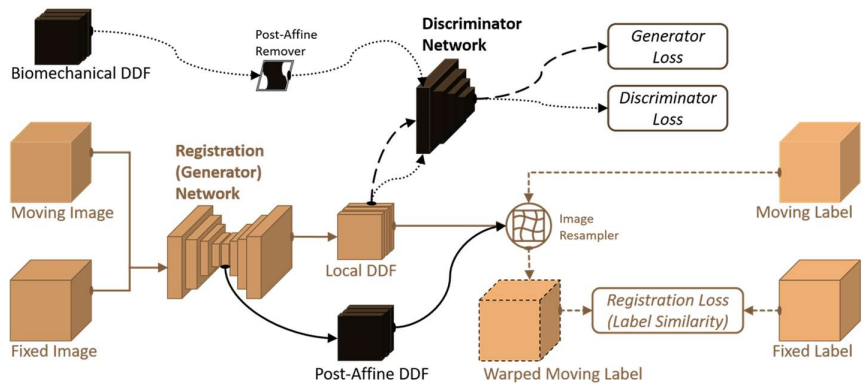

**Fig. 1.** The lighter shaded components connected by straight lines illustrate the weakly-supervised network training for multimodal image fusion [4]. The darker shaded components connected by curved lines depict the added elements that enable the proposed adversarial deformation regularization. Data flows required during inference, i.e. registration, are connected by solid lines, while other data connected by dotted- or dashed lines are only required for training.

### 2.2 Registration Loss for Weakly Supervised Multimodal Image Fusion

Assume $N$ pairs of moving- and fixed images for training, $\{\mathbf{x}_n^A\}$ and $\{\mathbf{x}_n^B\}$, respectively, $n = 1, \dots, N$. Further, assume $M_n$ pairs of moving- and fixed labels, $\{\mathbf{y}_{mn}^A\}$ and $\{\mathbf{y}_{mn}^B\}$, representing corresponding anatomical structures identified in the $n^{th}$ image pair, $m = 1, \dots, M_n$. The training of a registration network aims to predict inverse DDF $\boldsymbol{u}^{reg}$ that minimizes a negative *expected label similarity* over $N$ training image pairs:

$$\mathcal{L}^{(reg)} = -\frac{1}{N}\sum_{n=1}^{N}\frac{1}{M_n}\sum_{m=1}^{M_n} J_{mn}\big(\mathcal{T}\big(\mathbf{y}_{mn}^A, \mathbf{u}_n^{reg}\big), \mathbf{y}_{mn}^B\big) \qquad (1)$$

where the inner summation represents the image-level label similarity, averaging a label-level similarity measure $J_{mn}$ over $M_n$ labels associated with the $n^{th}$ image pair. Given a network-predicted displacement field $\mathbf{u}_n^{reg}(\mathbf{x}_n^A, \mathbf{x}_n^B; \theta^{(reg)})$, the label-level similarity is computed between the fixed label $\mathbf{y}_{mn}^B$ and the spatially warped moving label $\mathcal{T}(\mathbf{y}_{mn}^A, \mathbf{u}_n^{reg})$. We adopt a differentiable, efficient and imaging-modality-independent multiscale-Dice $J_{mn} = \frac{1}{Z}\sum_\sigma S_{Dice}\left(f_\sigma\left(\mathcal{T}(\mathbf{y}_{mn}^A, \mathbf{u}_n^{reg})\right), f_\sigma(\mathbf{y}_{mn}^B)\right)$, where $f_\sigma$ is a 3D Gaussian filter with an isotropic standard deviation $\sigma$ (here, $\sigma \in \{0, 1, 2, 4, 8, 16, 32\}$ in mm and the number of scales $Z = 7$). $f_{\sigma=0}$ denotes unfiltered binary labels at the original scale included in averaging the soft probabilistic Dice values $S_{Dice}$. The moving- and fixed images are the only network inputs. Therefore, the subsequent inference, i.e. registration, does not require anatomical labels, as illustrated in Fig. 1.

Displacement fields predicted by the multimodal registration network comprise of two combined geometric transformations: the biophysical deformation (deformation of anatomical structures) which should be regularized by the biomechanical simulations and the imaging-coordinate-system changes which should not. The imaging-coordinate-system changes reflect case-specific intra-procedural state (ultrasound imaging parameters such as probe position, field-of-view relative to anatomy, 3D voxel calibration and reconstruction) that is needed for intra-procedural registration and is not present in the biomechanical simulations. Therefore, to decouple these, the proposed network generates two transformations: a local DDF $\mathbf{u}_n^{local}$ intended to model only the biophysical deformation and an affine DDF $\mathbf{u}_n^{global}$ intended to model coordinate-system changes, as illustrated in Fig. 1. In minimizing the registration loss $\mathcal{L}^{(reg)}$, these are composed and optimised jointly, i.e. $\mathcal{T}(\mathbf{y}_{mn}^A, \mathbf{u}_n^{reg}) = \mathcal{T}(\mathcal{T}(\mathbf{y}_{mn}^A, \mathbf{u}_n^{local}), \mathbf{u}_n^{global})$. To regularize the biophysical deformation alone, the network is trained such that the predicted local DDFs $\mathbf{u}_n^{local}$ match a regularizing data distribution (the FE-simulated data distribution described in Section 2.3) that has been normalized to exclude affine variation.

### 2.3 Adversarial Losses based on Biomechanical Simulations

From a separate patient data set, assume a total of $S$ FE simulations calculating the deformed nodal positions of the prostate glands and surrounding anatomical regions, defined on patient-specific tetrahedral meshes fitted to segmentations of the zonal structures, bladder, rectum and pelvic bones [5, 6]. For each simulation, the nonlinear neo-Hookean material properties of different regions and the boundary conditions, including initial position and movement of a virtual TRUS probe with variable-sized acoustic coupling balloon, are randomly sampled to cover the variance in intra-procedural scenarios. Inverting simulated deformation fields $\mathbf{v}_s^{sim}$ maps the deformed FE nodes $\mathbf{y}_s^1$ back to the undeformed $\mathbf{y}_s^0$, such that $\mathbf{y}_s^0 = \mathcal{T}^{-1}(\mathbf{y}_s^1, \mathbf{v}_s^{sim}), s = 1, \dots, S$.

To normalize the data distribution in deformation space, each $\mathbf{v}_s^{sim}$ is decomposed into a global affine transformation $\mathbf{v}_s^{global}$ and an affine-removed local inverse displacement field $\mathbf{v}_s^{local}$, such that $\mathbf{y}_s^0 = \mathcal{T}^{-1}(\mathcal{T}^{-1}(\mathbf{y}_s^1, \mathbf{v}_s^{global}), \mathbf{v}_s^{local})$. Using a linear least-squares method, $\mathbf{v}_s^{global}$ are computed to minimize $\|\mathcal{T}^{-1}(\mathbf{y}_s^1, \mathbf{v}_s^{global}) - \mathbf{y}_s^0\|^2$ before training. While the predicted- and simulated global transformations may have different distributions (as discussed in Section 2.2), the local transformations should have the same distribution. Specifically, the distribution of registration-predicted local DDFs $P_{reg}$, represented by random vector $\mathbf{u}^{local} \sim P_{reg}$ with samples $\{\mathbf{u}_n^{local}\}$, can be regularized by comparing to the FE-simulated data distribution $P_{sim}$, represented by random vector $\mathbf{v}^{local} \sim P_{sim}$ with samples $\{\mathbf{v}_s^{local}\}$. In this work, we adopt a stable discriminator loss $\mathcal{L}^{(dis)}$ and a non-saturating generator loss $\mathcal{L}^{(gen)}$ based on Jensen-Shannon divergence [10],

$$\mathcal{L}^{(dis)} = -\frac{1}{2}\mathbb{E}_{\mathbf{v}^{local}} \log D(\mathbf{v}^{local}) - \frac{1}{2}\mathbb{E}_{\mathbf{u}^{local}} \log\left(1 - D(\mathbf{u}^{local})\right) + \frac{\gamma}{2}\Omega(\mathbf{u}^{local}, \mathbf{v}^{local}) \qquad (2)$$

and

$$\mathcal{L}^{(gen)} = -\frac{1}{2}\mathbb{E}_{\boldsymbol{u}^{local}} \log D(\boldsymbol{u}^{local}) \qquad (3)$$

respectively, where $\mathbb{E}$ denotes statistical expectation. A distribution smoothing term $\Omega(\boldsymbol{u}^{local}, \boldsymbol{v}^{local})$ is added to stabilize adversarial training [10], weighted by an annealing scalar $\gamma$ (here, exponentially decaying from 0.2 to 0.05 for the normalized data described in Section 3). Importantly, this annealing regularization also has a favorable effect on the registration network that encourages the affine branch to learn global transformation, so that the trained local DDF contains minimum affine component. It may be because that the smoothed distribution back-propagates stronger gradients from the generator loss to the local-DDF branch, relatively dominating its registration loss, especially during initial training stage when $\gamma$ is large.

Without loss of generality, displacement samples from MR-, TRUS- and FE coordinates have differently truncated finite sampling domains. As TRUS images have the most restricted fields-of-view, MR and TRUS are considered as moving- and fixed images, respectively, when computing registration loss. To avoid sampling larger-domain (MR and FE in this case) displacements from smaller-domain (TRUS) when computing adversarial losses, each FE-simulated DDF $\boldsymbol{v}_s^{local}$ is resampled from an estimated TRUS field-of-view before removing the affine component. The resampling coordinates are determined by matching the bounding boxes of the *deformed* prostate glands in FE- and TRUS coordinates, the latter of which is randomly sampled from the training TRUS data, i.e. the fixed labels. This estimate is also aided by data augmentation described in Section 2.4 to represent the variation in sampling domain.

## 2.4 Network Architectures and Training

As shown in Fig. 2, the registration network adapts a 3D encoder-decoder architecture taking a concatenated image pair as input, down-sampled and up-sampled by convolution (conv) and transpose-convolution (deconv), respectively, both with strides of two. The encoder consists of four residual network (resnet) blocks using 3×3×3 conv kernels, with increasing numbers of feature channels $n_{0-4}$ and decreasing feature map sizes $s_{0-4}$, both by a factor of two. The decoder has four *reverse* resnet blocks with, additionally, four trilinear additive up-sampling layers added over the deconv layers. Four summation skip layers shortcut the network resolution levels. Five trilinear-up-sampled displacement summands $\delta_{0-4}$ across levels $s_{0-4}$ are summed to predict the output local DDF. 12 output affine parameters were predicted by an additional resnet block, branched out from the deepest encoder layer $s_4$. The discriminator shares a similar architecture with the registration network encoder, with first layer batch normalization (BN) removed and rectified linear units (relu) replaced by leaky relu (lrelu) [11]. It accepts input DDF x-, y- and z-channels and predicts binary classification logits after a fully-connected projection. Both networks start with $n_0$=32 initial channels.

The networks were implemented in TensorFlow™ with open-source code from NiftyNet [12]. For data augmentation, each image-label pair was warped by a random affine transformation and each simulated DDF was composed with a random affine for varying the sampling domain (as discussed in Section 2.3), before being fed into training. Using the Adam optimizer starting at a learning rate of $10^{-6}$ for both registration- and discriminator networks, each model was trained for 36 hours with a minibatch size of 4 on a 24GB NVIDIA® Quadro™ P6000 GPU card. The adversarial weight $\lambda_{adv}$ was set to 0.01 for the reported results.

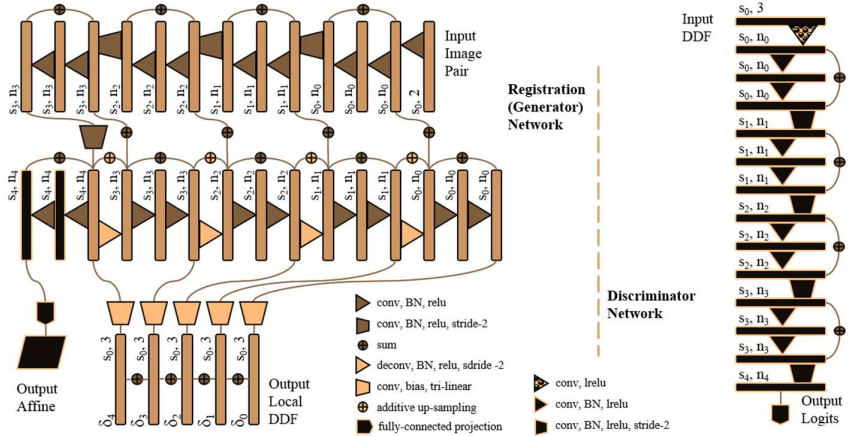

**Fig. 2.** The proposed registration- and discriminator networks (see details in Section 2.4).

## 3  Experiments and Results

For computing the registration loss, a total of 108 pairs of T2-weighted MR- and TRUS images from 76 patients were acquired in multiple biopsy or therapy (ClinicalTrials.gov Identifiers: NCT02290561, NCT02341677) clinical trials. Using a clinical ultrasound machine with a transperineal probe, 57-112 sagittal TRUS frames were acquired for each patient by rotating a digital brachytherapy stepper to reconstruct 3D volumes in Cartesian coordinates. Both MR- and TRUS volumes were normalized to zero-mean with unit-variance intensities after being resampled to 0.8×0.8×0.8 mm$^3$ voxels. For assessing the regularization efficacy, gland segmentations were used as the only type of training landmarks, i.e. $M_n$=1 in Eq. (1), which are arguably the most easy-to-annotate landmarks for both imaging modalities with many automated algorithms [13]. Gland segmentations on MR were acquired as per the trial protocols and those on TRUS were contoured on original slices. Both gland masks were then resampled to the voxel sizes of the associated MR or TRUS. For the adversarial training, MR images for FE meshing were acquired from an independent group of 143 patients who underwent the same procedures, without using their TRUS data in this study. For each patient, 500 FE simulations required 3-4 GPU-hours using a nonlinear FE solver [14]. Both the simulated- and predicted DDFs, as inputs of the discriminator, are normalized such that the simulated data have zero-mean and unit-variance displacements.

For quantitative validation, a total of 834 pairs of corresponding anatomical landmarks from the 108 paired images were manually labelled and further verified/edited by second observers, including apex, base, urethra, gland zonal separations, visible lesions, junctions between gland, vas deference and seminal vesicles, and other *ad hoc* landmarks such as calcifications and cysts. The annotation process took more than two hundred man-hours. Based on these independent validation landmarks, the proposed adversarial regularization was compared with two widely used smoothness regularizers, by adding a weighted $L^2$-norm of displacement gradients or bending energy to the registration loss in Eq. (1). For the reported results, both weights were set to 0.5, which produced the lowest median TREs from eight cross-validation experiments with four different weighting values, 0.01, 0.1, 0.5 and 1. In each fold of the 12-fold patient-level cross-validation experiments, 6-7 test patients were held out while the data from the remainder patients were used in training with all 71,500 FE simulations. The TRE was defined as root-mean-square centroid distance between the warped- and fixed validation labels. Dice similarity coefficient (DSC) was computed between the binary gland masks. These two test data results reflect quantitative clinical requirements in localizing target anatomy such as MR-visible tumors and avoiding heathy surroundings.

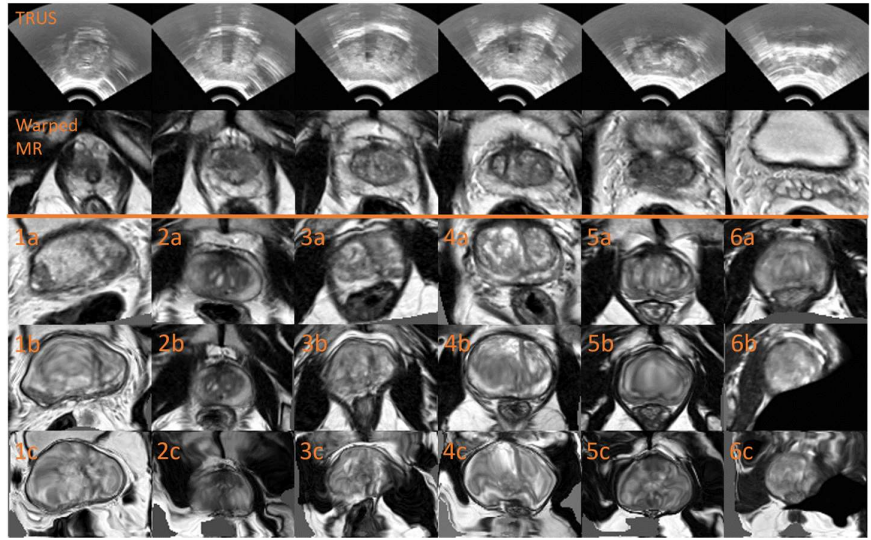

**Fig. 3.** The first two rows show example slices from a TRUS volume and the MR volume registered using the proposed network. The example warped MR slices (at the same slice locations in each patient, 1 - 6) are also compared between the proposed adversarial regularization (a), the bending energy (b) and registration without smoothness penalty (c).

**Table 1.** Medians [1st quatiles, 3rd quatiles] of TRE and DSC results from cross-validation

|  | **Adversarial** | **Bending Energy** | **$L^2$-norm** | **No Regularization** |
| --- | --- | --- | --- | --- |
| **TRE** | 6.3 [3.4, 8.7] | 9.5 [4.6, 13.0] | 10.2 [5.1, 14.7] | 16.3 [14.1, 23.8] |
| **DSC** | 0.82 [0.76, 0.87] | 0.90 [0.83, 0.92] | 0.91 [0.84, 0.92] | 0.93 [0.88, 0.95] |

Approximately four 3D automatic registrations per second can be performed on the same GPU, which is adequate for many interventional applications. Fig. 3 contains example registered images using the proposed network. The adversarial regularization appears more likely to preserve local details and, most interestingly, generates motion patterns unseen in those with other regularization, e.g. near-rigid motion around the rectum area where the virtual ultrasound probe is placed in FE simulations. As summarized in Table 1, the adversarial regularized registrations produced a significantly lower median TRE than the networks trained with $L^2$-norm or bending energy did (both *p-values<0.001*, paired Wilcoxon signed-rank tests at *α=0.05*), consistent with the visual inspection. The higher DSCs with $L^2$-norm, bending energy or without regularization may therefore indicate overfitting to the training gland labels. The obtained TRE results were based on 108 image pairs, compared to 8-19 patients validated in several previous work [5-7]. These still seem to be higher than that of 4.2 mm reported in [4], in which 4,000 training labels were required. Further comparisons, such as a comprehensive sampling of hyper-parameter values, may conclusively quantify the adversarial regularization, such as the trade-off of accuracy when using more training landmarks.

## 4   Conclusion

In this work, we have proposed a novel adversarial deformation regularization, a potentially versatile strategy incorporating model-based constraints to assist data-driven image registration algorithms. We report promising results based on validation on a substantial interventional imaging data set from prostate cancer patients. Potential for further improving registration performance by, for instance, leveraging between the requirements of anatomical labels, universal smoothness measures and the proposed adversarial priors may be of interest in future research and clinical adoption.

10. Roth, K. et al., 2017, Stabilizing training of generative adversarial networks through regularization. *NIPS 2017*, 2015-2025.
11. Radford, A. et al., 2015, Unsupervised representation learning with deep convolutional generative adversarial networks. *arXiv preprint* arXiv:1511.06434.
12. Gibson, E. et al., 2018, NiftyNet: a deep-learning platform for medical imaging. *Computer Methods & Programs in Biomedicine*, 158, 113-122.
13. Litjens, G. et al., 2014, Evaluation of prostate segmentation algorithms for MRI: the PROMISE12 challenge." *Medical image analysis*, 18(2): 359-373.
14. Johnsen, S.F. et al., 2015, NiftySim: A GPU-based nonlinear finite element package for simulation of soft tissue biomechanics. *IJCARS*, 10(7): 1077-1095.